\documentclass[conference]{IEEEtran}
\IEEEoverridecommandlockouts
\usepackage{cite}
\usepackage{amsmath,amssymb,amsfonts}
\usepackage{algorithmic}
\usepackage{graphicx}
\usepackage{textcomp}
\usepackage{xcolor}
\usepackage{subcaption}
\usepackage{multirow}
\usepackage{threeparttable}

\def\BibTeX{{\rm B\kern-.05em{\sc i\kern-.025em b}\kern-.08em
    T\kern-.1667em\lower.7ex\hbox{E}\kern-.125emX}}
\begin{document}

\title{Discriminative Pedestrian Features and Gated Channel Attention for Clothes-Changing Person Re-Identification\\
\thanks{This work was supported in part by the National Natural Science Foundation of China under Grant 62172212, in part by the Natural Science Foundation of Jiangsu Province under Grant BK20230031. *Liyan Zhang is the corresponding author.}
}

\author{
\IEEEauthorblockN{1\textsuperscript{st} Yongkang Ding}
\IEEEauthorblockA{\textit{College of Computer Science and Technology} \\
\textit{Nanjing University of Aeronautics and Astronautics}\\
Nanjing, China \\
ykding@nuaa.edu.cn} \\

\IEEEauthorblockN{3\textsuperscript{rd} Hanyue Zhu}
\IEEEauthorblockA{\textit{College of Computer Science and Technology} \\
\textit{Nanjing University of Aeronautics and Astronautics}\\
Nanjing, China \\
zhnyue@nuaa.edu.cn} \\

\IEEEauthorblockN{5\textsuperscript{th} Liyan Zhang*}
\IEEEauthorblockA{\textit{College of Computer Science and Technology} \\
\textit{Nanjing University of Aeronautics and Astronautics}\\
Nanjing, China \\
zhangliyan@nuaa.edu.cn} 
\and
\IEEEauthorblockN{2\textsuperscript{nd} Rui Mao}
\IEEEauthorblockA{\textit{School of Information and Artificial Intelligence} \\
\textit{Anhui Agriculture University}\\
Hefei, China \\
22723005@stu.ahau.edu.cn} \\

\IEEEauthorblockN{4\textsuperscript{th} Anqi Wang}
\IEEEauthorblockA{\textit{College of Computer Science and Technology} \\
\textit{Nanjing University of Aeronautics and Astronautics}\\
Nanjing, China \\
aqwang@nuaa.edu.cn} 

}

\maketitle

\begin{abstract}
In public safety and social life, the task of Clothes-Changing Person Re-Identification (CC-ReID) has become increasingly significant. However, this task faces considerable challenges due to appearance changes caused by clothing alterations. Addressing this issue, this paper proposes an innovative method for disentangled feature extraction, effectively extracting discriminative features from pedestrian images that are invariant to clothing. This method leverages pedestrian parsing techniques to identify and retain features closely associated with individual identity while disregarding the variable nature of clothing attributes. Furthermore, this study introduces a gated channel attention mechanism, which, by adjusting the network's focus, aids the model in more effectively learning and emphasizing features critical for pedestrian identity recognition. Extensive experiments conducted on two standard CC-ReID datasets validate the effectiveness of the proposed approach, with performance surpassing current leading solutions. The Top-1 accuracy under clothing change scenarios on the PRCC and VC-Clothes datasets reached 64.8\% and 83.7\%, respectively.
\end{abstract}


\section{Introduction}
With the rapid development of computer vision\cite{tang-1,tang-2}, person re-identification (ReID) has emerged as a highly focused research area. Pedestrian ReID aims to connect the movement trajectories of target individuals across different surveillance areas, facilitating pedestrian tracking across time, locations, and devices. For large-scale video surveillance systems, ReID technology enhances the efficiency and accuracy of pedestrian retrieval, compensating for the visual limitations of fixed cameras, and holds significant application value in smart cities, surveillance security, judicial investigation, and pandemic prevention and control. However, a prominent challenge faced by ReID today is the issue of cloth-changing person re-identification (CC-ReID)\cite{my-1,my-2}.\par

\begin{figure}[htbp]
\centerline{\includegraphics[width=0.5\textwidth]{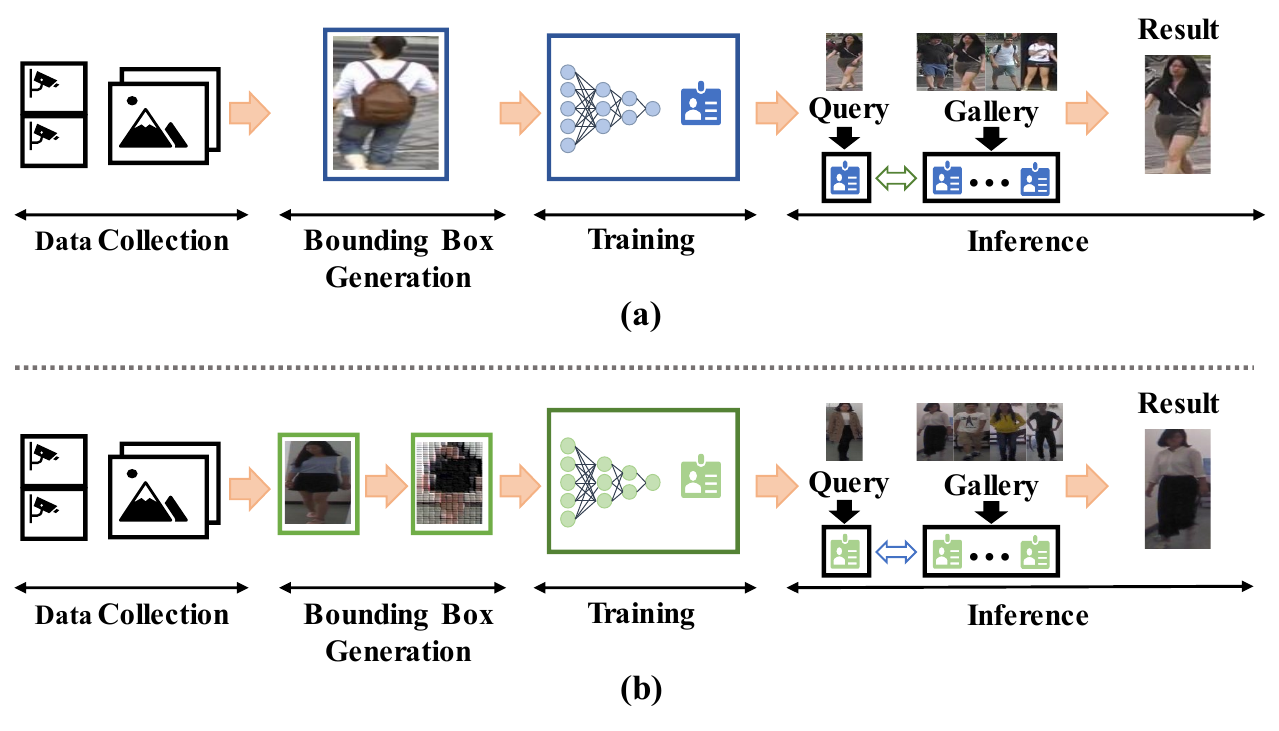}}
\caption{Flowchart of the ReID process. (a) Illustration of conventional ReID
model. (b) Illustration of our CC-ReID model.}
\label{fig1}
\end{figure}

CC-ReID refers to the difficulty in accurately identifying individuals across different times or locations due to changes in attire. For instance, in video surveillance, a pedestrian might change their coat, wear a hat, or don a mask, leading to substantial alterations in their appearance and, consequently, increasing the complexity of ReID tasks. Traditional ReID methods typically rely on appearance features such as color, texture, and shape. However, these methods often fall short in tackling the CC-ReID challenge, as changes in attire can significantly increase the disparity between feature vectors, thereby reducing matching accuracy. Figure ~\ref{fig1} illustrates the distinction between CC-ReID and classical ReID tasks.

In recent years, the swift advancement of deep learning technologies has opened new avenues for addressing the CC-ReID problem. Deep learning models, by learning more representative feature representations, have enhanced the performance of ReID. Significant progress has been made by researchers employing deep learning models like convolutional neural networks and Transformers. These models are capable of learning a richer set of pedestrian features, including local details, posture information, and semantic representations, thus improving the accuracy and robustness of ReID. Existing solutions to this challenge can be categorized into two main types:

\noindent
\textbf{\emph{(i)}} One approach involves the use of multimodal methods to incorporate additional information (such as sketches\cite{prcc}, 3D shapes\cite{3d}, gait data\cite{gait}, and skeletal\cite{ltcc} structures) to learn features that are independent of clothing. However, this may overlook critical intrinsic human details like facial expressions, arm postures, and leg shapes. These overlooked aspects play a key role in pedestrian re-identification, providing richer distinguishing features.

\noindent
\textbf{\emph{(ii)}} The other approach, based on Generative Adversarial Networks (GANs), is employed to learn the variations in a pedestrian's appearance over time. However, this method inevitably alters the original RGB images, introducing unrealistic or distorted details. This could lead to model instability and training failures, requiring extended training durations and meticulous parameter tuning.\par

\begin{figure*}[htbp]
\centerline{\includegraphics[width=0.9\textwidth]{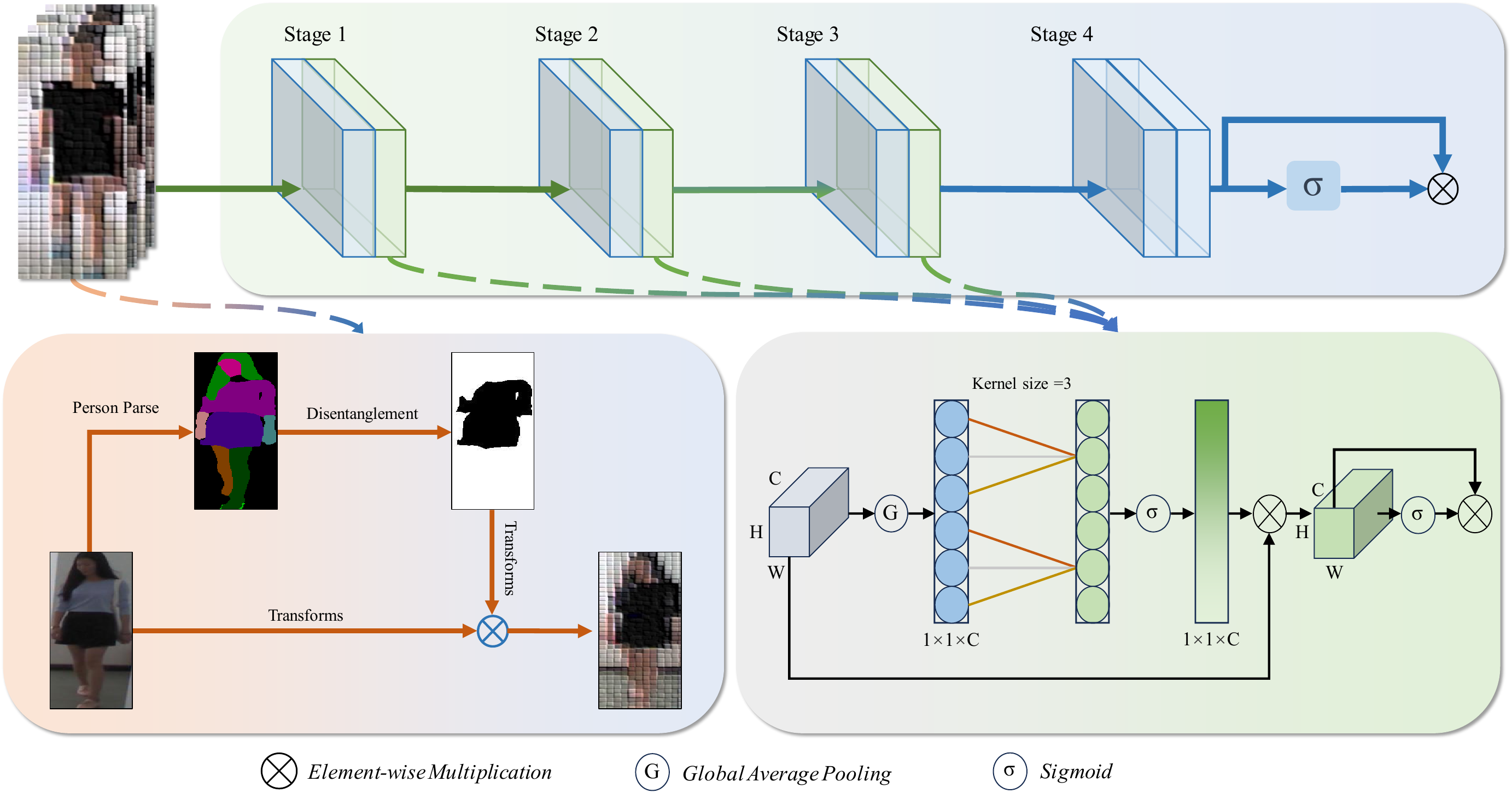}}
\caption{Pipeline of the Proposed Method. By extracting pedestrian discriminative features from semantic segmentation maps and incorporating a gated channel attention mechanism into the network, the performance of the model is further enhanced.}
\label{fig2}
\end{figure*}

In the process of exploring solutions to issues present in existing methods, we propose a novel Disentangled Gated Attention Mechanism model. The core innovation of this model lies in the application of a disentangled approach to extract clothing-irrelevant features. By utilizing pedestrian parsing and controlled disentanglement techniques, we extract clothing-irrelevant semantic segmentation features from RGB images, which are then transformed into dual-precision feature maps. These maps are fused with the original RGB images to extract distinctive pedestrian features. Building upon this, to further enhance the model's expressiveness and generalization capability, we introduce a Gated Channel Attention Mechanism based on the existing channel attention\cite{eca} framework. This mechanism initially employs the channel attention mechanism to learn and analyze the importance of each feature channel, assigning different weight coefficients to each channel. This effectively enhances key features and suppresses irrelevant ones. More importantly, by integrating the gating mechanism, the model selectively focuses on and utilizes crucial information within the input features, while filtering out noise or secondary information.

In this study, we adopted a dual-phase optimization strategy for training the model. During the initial $n$ epochs, the model learns solely through identity loss, acquiring fundamental skills for CC-ReID. Subsequently, triplet loss is integrated, working in tandem with identity loss to enhance the model's performance in complex CC-ReID tasks. This approach amplifies the model's capability to handle intricate scenarios in the field of computer vision. 

The contributions of this article can be summarized as follows:

\begin{itemize}
    \item We have developed a controllable disentangled feature extraction method that extracts pedestrian identity features independent of clothing. This feature is also used in the inference stage for final identity prediction. 
    \item A gated channel attention mechanism has been integrated into our model to enhance the representation of important features and suppress less significant ones, thereby improving the model's expressive capability.
    \item We adopted an innovative dual-stage optimization strategy in the training phase. This approach is aimed at enhancing the model's foundational learning capabilities, as well as improving its ability to handle complex scenarios.
    \item We conducted extensive experiments on both a real-world and a virtual-world dataset, demonstrating that our method outperforms the current state-of-the-art approaches in the task of CC-ReID.
\end{itemize}

\section{The Proposed Method}
The proposed method employs stages 1 to 4 of ResNet\cite{resnet} as its backbone network for feature extraction. Figure 2 illustrates the structure of this method. During the feature extraction process, a controllable disentanglement approach is used to extract pedestrian identity features that are independent of clothing. Additionally, a gated channel attention mechanism is introduced in stages 1 to 3 of ResNet, while only the gating mechanism is used in stage 4, aiming to enhance the overall expressive capability of the model.

\subsection{Controllable Disentanglement Feature Extraction Method (CDM)}

In the context of feature extraction, the controllable disentanglement method refers to an approach where we actively eliminate the interference caused by clothing variations in order to extract discriminative pedestrian features. This method aims to force the model to learn features that are independent of clothing changes. The process, as illustrated in the bottom left corner of Figure ~\ref{fig2}, begins with pedestrian parsing on the RGB image to generate a semantic segmentation map. Subsequently, the controllable disentanglement method is applied to transform the semantic segmentation map into a disentangled grayscale image that is not influenced by clothing. The pixel values in this grayscale image range from 0 (black) to 1 (white). Following this, a series of transformations are applied to obtain a disentangled grayscale feature map that is independent of clothing. To capture more semantic information, this grayscale feature map is fused with the corresponding feature map of the RGB image, achieving the objectives of the controllable disentanglement method.

We used the SCHP\cite{schp} model, pre-trained on the LIP\cite{lip} dataset, for pedestrian parsing of our dataset. Each image in the LIP dataset contains major human body parts, clothing information, and background, represented with pixel-level annotations. By applying the SCHP model to our dataset, we obtain semantic segmentation annotations for each image, featuring 20 body part categories. To extract the clothing-independent feature map, we extracted the pixel values of the following parts from the semantic segmentation map: ['Background', 'Hair', 'Face', 'Left-arm', 'Right-arm', 'Left-leg', 'Right-leg', 'Left-shoe', 'Right-shoe']. Subsequently, the semantic segmentation map is transformed into a dual-precision grayscale image, facilitating subsequent feature fusion operations.\par

\subsection{Gated Channel Attention Mechanism (GCA)}

In this study, we propose an innovative gated channel attention mechanism designed to enhance the performance of the ResNet50 model in the task of pedestrian re-identification. The core idea of this mechanism is to finely regulate the importance of feature channels, allowing the model to more effectively focus on the features crucial for discerning pedestrian identities. Our mechanism is implemented in different stages of the ResNet network to adapt to the characteristics of features at each stage.\par

In the first three stages of ResNet (i.e., stage 1 to stage 3), we integrate both channel attention and gating mechanisms. Specifically, the channel attention component employs an effective channel\cite{eca} attention model that dynamically adjusts channel weights by considering local interactions between channels. This enables the network to pay more attention to features helpful for the current task, such as facial, postural, and arm features. Concretely, the channel attention module first performs global average pooling on the input feature map $X$, followed by one-dimensional convolution and Sigmoid activation to obtain channel weights $\omega$. Subsequently, the original input feature map is element-wise multiplied with the weights to obtain the final output feature map $A$, expressed as:

\begin{equation}
\omega=\sigma(Conv_{1D}(Avgpool(X))),\label{eq1}
\end{equation}

\begin{equation}
A=X\odot \omega,\label{eq2}
\end{equation}

\noindent
Simultaneously, the gating mechanism achieves self-regulation based on the features themselves. Specifically, the feature \(A\) first undergoes a sigmoid activation function, obtaining a weight between 0 and 1. Subsequently, this weight is multiplied with the original feature \(A\), realizing self-regulation of the features. This combined strategy allows the model to emphasize crucial channels while maintaining effective balance among features during the extraction of pedestrian features. The formula expression is as follows, where \(G\) represents the features fed into the next stage:

\begin{equation}
G=A\odot\sigma(A).\label{eq3}
\end{equation}

For the fourth stage (stage4) of ResNet, we have chosen to exclusively employ a gating mechanism. This decision is grounded in a thorough understanding of the characteristic representations at different stages of the network. At this deeper stage of the network, features have become more abstract and compound. Direct channel recalibration could lead to excessive distortion of features. Hence, we utilize a gating mechanism to individually adjust the features, thereby maintaining the integrity of higher-level features. Such a design endows the model with enhanced robustness and accuracy, particularly evident in the complex task of recognizing pedestrians during clothing changes.

\subsection{Training Loss}
In this study, we have implemented a two-stage optimization strategy for training our model for re-identification of pedestrians undergoing clothing changes. This approach not only underscores the importance of basic feature learning but also aims to enhance the model's capability in handling complex scenarios. During the initial 'n' epochs of model training, we focus on employing identity loss to train the model. The essence of this stage lies in enabling the model to effectively recognize pedestrians wearing identical clothing. Identity loss facilitates the learning of fundamental characteristics of individuals, such as body shape, posture, and facial features. The objective of this phase is to establish a robust foundation, enabling the model to adapt more effectively to complex tasks in subsequent learning stages. Identity loss is represented as follows: 

\begin{equation}
\begin{aligned}
\mathcal{L}_{ID}=-\sum_{i=1}x_ilogy_i,
\end{aligned}
\end{equation}

\noindent
where \( x_i \) represents the \( i \)-th element of the true label, and \( y_i \) denotes the probability predicted by the model for \( x \) belonging to the \( i \)-th category.

Upon entering the 'n-th' epoch, we introduce the Triplet Loss, which works in conjunction with the identity loss on the model. The inclusion of Triplet Loss\cite{triplet} is aimed at enhancing the model's ability to differentiate pedestrians wearing different clothes. In this stage, the model is expected not only to identify distinct individuals but also to discern those with similar features even when they are wearing different attire. This combined loss strategy effectively augments the model's capability in handling hard-to-recognize samples, especially under varying clothing conditions. The metric function is represented using the Euclidean distance, and the formula is as follows:

\begin{equation}
\begin{aligned}
d_{pos}=\|E_{ap}(x_{anc})-E_{ap}(x_{pos})\|_2,
\end{aligned}
\end{equation}

\begin{equation}
\begin{aligned}
d_{neg}=\|E_{ap}(x_{anc})-E_{ap}(x_{neg})\|_2,
\end{aligned}
\end{equation}

\begin{equation}
\begin{aligned}
\mathcal{L}_{Tri}=[\alpha+d_{pos}-d_{neg} , 0]_+,
\end{aligned}
\end{equation}

\noindent
where $x_{anc}$ denotes the given anchor image. $x_{pos}$ is considered as the positive sample sharing the same ID as $x_{anc}$, and $x_{neg}$ is the negative sample from a different ID. The hyperparameter $\alpha$ is utilized to regulate the differential distance between the pairs of positive and negative samples.

This dual-stage training methodology enables our model to exhibit superior performance in the task of re-identifying pedestrians undergoing clothing changes. This approach not only improves the model's basic recognition capabilities but also enhances its robustness and accuracy in dealing with complex, diverse scenarios. The overall loss function for the two-stage approach is depicted as follows:

\begin{equation}
\begin{aligned}
\mathcal L_{\text {1 }}= \mathcal L_{ID}
\end{aligned},
\end{equation}

\begin{equation}
\begin{aligned}
\mathcal L_{\text {2 }}= \lambda_{1} \mathcal L_{ID}+ \lambda_{2} \mathcal L_{Tri}
\end{aligned},
\end{equation}

\noindent
where $\mathcal L_{ID}$ represents the identity loss, and $\mathcal L_{Tri}$ denotes the triplet loss, with $\lambda_{1}$ and $\lambda_{2}$ serving as the hyper-parameter to balance these two losses. Finally, the network model is optimized by minimizing $\mathcal L_{\text {1 }}$ and $\mathcal L_{\text {2 }}$, respectively.

\section{Experiments}

\begin{table*}[htbp]
\renewcommand{\arraystretch}{1.1}
\renewcommand\tabcolsep{10pt}
\centering
\caption{\centering Comparison With State-of-the-art Methods On PRCC and VC-Clothes.}\label{tab1}
\begin{threeparttable}
    \begin{tabular}{c|c|cccc|cccc}
\hline
\multirow{3}{*}{Method} & \multirow{3}{*}{Venue} & \multicolumn{4}{c|}{PRCC}                                                         & \multicolumn{4}{c}{VC-Clothes}                                                     \\ \cline{3-10} 
                        &                        & \multicolumn{2}{c|}{Same Cloth}                           & \multicolumn{2}{c|}{Clothing Change}       & \multicolumn{2}{c|}{Same Cloth}                            & \multicolumn{2}{c}{Clothing Change}        \\ \cline{3-10} 
                        &                        & Top-1        & \multicolumn{1}{c|}{mAP}           & Top-1         & mAP           & Top-1         & \multicolumn{1}{c|}{mAP}           & Top-1         & mAP           \\ \hline
PCB\cite{pcb}                     & ECCV 18                & 99.8         & \multicolumn{1}{c|}{97.0}          & 41.8          & 38.7          & 94.7          & \multicolumn{1}{c|}{94.3}          & 62.0          & 62.3          \\
IANet\cite{ianet}                   & CVPR 19                & 99.4         & \multicolumn{1}{c|}{98.3}          & 46.3          & 45.9          & -             & \multicolumn{1}{c|}{-}             & -             & -             \\
ISP\cite{isp}                     & ECCV 20                & 92.8         & \multicolumn{1}{c|}{-}             & 36.6          & -             & 94.5          & \multicolumn{1}{c|}{94.7}          & 72.0          & 72.1          \\
TransReID\cite{transreid}               & ICCV 21                & 97.3         & \multicolumn{1}{c|}{95.9}          & 47.1          & 49.3          & \textbf{95.1} & \multicolumn{1}{c|}{94.5}          & 70.0          & 71.8          \\ \hline
3DSL\cite{3d}                    & CVPR 21                & -            & \multicolumn{1}{c|}{-}             & 51.3          & -             & -             & \multicolumn{1}{c|}{-}             & 79.9          & 81.2          \\
FSAM\cite{fsam}                    & CVPR 21                & 98.8         & \multicolumn{1}{c|}{-}             & 54.5          & -             & 94.7          & \multicolumn{1}{c|}{94.8}          & 78.6          & 78.9          \\
GI-ReID\cite{gait}                 & CVPR 22                & 80.0         & \multicolumn{1}{c|}{-}             & 30.3          & -             & -             & \multicolumn{1}{c|}{-}             & 57.8          & 64.5          \\
CAL\cite{cal}                     & CVPR 22                & \textbf{100} & \multicolumn{1}{c|}{\textbf{99.8}} & 55.2          & 55.8          & \textbf{95.1} & \multicolumn{1}{c|}{\textbf{95.3}} & 81.4          & 81.7          \\
IMS-GEP\cite{ims-gep}                 & TMM 23                 & 99.7         & \multicolumn{1}{c|}{\textbf{99.8}} & 57.3          & \textbf{65.8}          & 94.7          & \multicolumn{1}{c|}{94.9}          & 81.8          & 81.7          \\ \hline
ours                    & -                      & 99.3         & \multicolumn{1}{c|}{94.3}          & \textbf{64.8} & 61.3 & 93.1          & \multicolumn{1}{c|}{92.8}          & \textbf{83.7} & \textbf{82.7} \\ \hline
\end{tabular}
    \begin{tablenotes}\footnotesize
        \item[] Note: Same Cloth ( SC ): Refers to computations performed using samples where the pedestrian's clothing is consistent; Clothing Change ( CC ): Denotes computations using samples with inconsistencies in pedestrian clothing.
    \end{tablenotes}
\end{threeparttable}
\end{table*}

\subsection{Datasets and Evaluation}
% \textbf{Datasets}
Our method was evaluated on two standard CC-ReID datasets: PRCC\cite{prcc} and VC-Clothes\cite{vc}. The PRCC dataset comprises 33,698 images from 221 individuals, each with two sets of clothing. VC-Clothes, a synthetic dataset derived from the game GTA5, consists of 19,060 photographs from 512 identities, captured by four cameras, with each identity wearing 1-3 different outfits. To assess our approach, we employed Top-1 accuracy and mean Average Precision (mAP). \par

% \textbf{Evaluations}
% We employed Top-1 accuracy and Mean Average Precision (mAP) to evaluate our method. Two testing setups were defined: (i) Clothing Change setting (CC), where we exclusively utilized samples with inconsistent clothing of pedestrians for the final computation; (ii) Same Clothing setting (SC), employing real samples with consistent clothing of pedestrians to calculate the final accuracy.

\subsection{Implement Details}
We employ ResNet50\cite{resnet} as the backbone network, setting the stride of the final convolution and down-sampling layer to 1. Simultaneously, the last average pooling is replaced with max-average pooling. Additionally, we initialize the backbone network using a pre-trained model from ImageNet\cite{imagenet}. Images were resized to 384 × 192 pixels. Data augmentation methods included random horizontal flipping, random cropping, and random erasing\cite{random}. During training, the Adam\cite{adam} optimizer was used for 120 epochs. The initial learning rate was set at 3.5$e^{-4}$, and it was reduced by a factor of 10 every 20 epochs. The batch size was configured to 32.\par
%  % For both PRCC and VC-Clothes datasets, the hyperparameter A was optimally set at 0.1 to achieve the best results.

\subsection{Comparison With the State-of-The-Art Methods}

Our proposed method was compared with the current state-of-the-art approaches on the PRCC and VC-Clothes datasets. The comparison included four classical ReID methods and five CC-ReID methods, as shown in Table ~\ref{tab1}. For pedestrian recognition in the same clothing scenario, our method was slightly inferior to some of the methods. This is attributed to the limitation of solely utilizing the disentangled features of pedestrians, resulting in the training process not capturing some clothing features of pedestrians without clothing changes. However, for the re-identification of pedestrians in clothing change scenarios, we achieved satisfactory results. In the clothing change scenarios of the PRCC and VC-Clothes datasets, our method attained a TOP-1/mAP of 64.8\%/61.3\% and 83.7\%/82.7\%, respectively. This is higher than the best-performing IMS-GEP method in Table ~\ref{tab1}, with an increase of 7.5\%/1.9\% in TOP-1 accuracy on the PRCC and VC-Clothes datasets.

\subsection{Ablation Experiments}
\subsubsection{Analysis of Different Components}
To evaluate the effectiveness of each component in our method, we conducted ablation studies on the PRCC dataset, as shown in Table ~\ref{tab2}. The experimental results indicate that using only the baseline yields higher performance in the same clothing setting but shows lower efficiency in clothing change scenarios. Notably, the introduction of the Controllable Disentanglement Feature Extraction Method (CDM) alone led to a significant performance improvement compared to the baseline model. This finding underscores the importance of CDM in enhancing the model's capabilities. Specifically, CDM effectively extracts and utilizes distinctive pedestrian features, thereby boosting the overall efficacy of the model. In contrast, the addition of the Gated Channel Attention Mechanism (GCA) alone did not result in a significant performance improvement, suggesting that GCA's contribution to model performance is limited when operating independently. However, when both the CDM and GCA modules were used simultaneously, we observed a performance improvement that exceeded the use of CDM alone, creating a synergistic effect. This synergy intensified the model's focus on distinctive pedestrian characteristics and improved its applicability in complex scenarios involving changes in pedestrian clothing. Furthermore, as shown in Table ~\ref{tab2}, relying on the CDM module, we employed a two-stage optimization strategy, which further enhanced the model's performance.

\begin{table}[htbp]
\renewcommand{\arraystretch}{1.1}
\renewcommand\tabcolsep{6pt}
\centering
\caption{\centering The ablation study of our method.}\label{tab2}
\begin{threeparttable}
    \begin{tabular}{cccc|cccc}
    \hline
    \multicolumn{4}{c|}{\multirow{2}{*}{method}} & \multicolumn{4}{c}{PRCC}                                                          \\ \cline{5-8} 
    \multicolumn{4}{c|}{}                        & \multicolumn{2}{c|}{SC}                           & \multicolumn{2}{c}{CC}        \\ \hline
    baseline      & CDM      & GCA      & *      & Top-1        & \multicolumn{1}{c|}{mAP}           & Top-1         & mAP           \\ \hline
    {\checkmark}             &          &          &        & \textbf{100} & \multicolumn{1}{c|}{99.3} & 53.3          & 53.5          \\
    {\checkmark}             & {\checkmark}        &          &        & 96.9         & \multicolumn{1}{c|}{90.5}          & 61.5          & 58.0          \\
    {\checkmark}             & {\checkmark}        &          & {\checkmark}      & 98.6         & \multicolumn{1}{c|}{93.6}          & 62.8          & 59.4          \\
    {\checkmark}             &          & {\checkmark}        &        & \textbf{100}          & \multicolumn{1}{c|}{\textbf{99.4}}          & 53.0          & 51.1          \\
    {\checkmark}             &          & {\checkmark}        & {\checkmark}      & \textbf{100}          & \multicolumn{1}{c|}{99.3}          & 53.0          & 51.1          \\
    {\checkmark}             & {\checkmark}        & {\checkmark}        &        & 98.0         & \multicolumn{1}{c|}{92.0}          & 63.9          & 60.0          \\
    {\checkmark}             & {\checkmark}        & {\checkmark}        & {\checkmark}      & 99.3         & \multicolumn{1}{c|}{94.3}          & \textbf{64.8} & \textbf{61.3} \\ \hline
    \end{tabular}
    \begin{tablenotes}\footnotesize
        \item[] Note: * indicates the use of a two-stage optimization process. Absence of this symbol signifies that the entire training process utilizes ID loss and triplet loss.
    \end{tablenotes}
\end{threeparttable}
\end{table}

\subsubsection{Analysis of Hyperparameter}
\begin{figure}[htbp]
    \centering
    \begin{subfigure}{0.43\linewidth}
        \includegraphics[width=\linewidth]{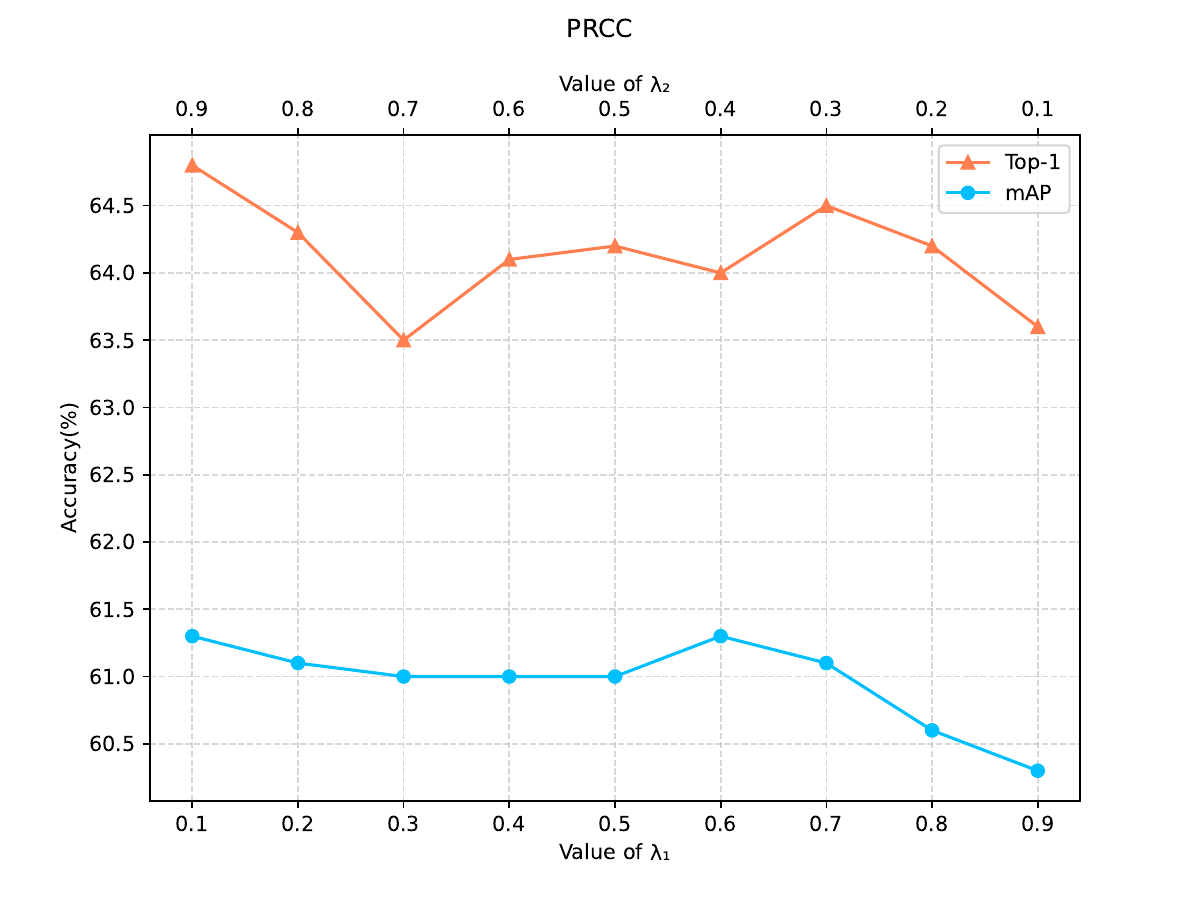}
        \caption{PRCC}
        \label{fig:sub1}
    \end{subfigure}
    \hfill
    \begin{subfigure}{0.43\linewidth}
        \includegraphics[width=\linewidth]{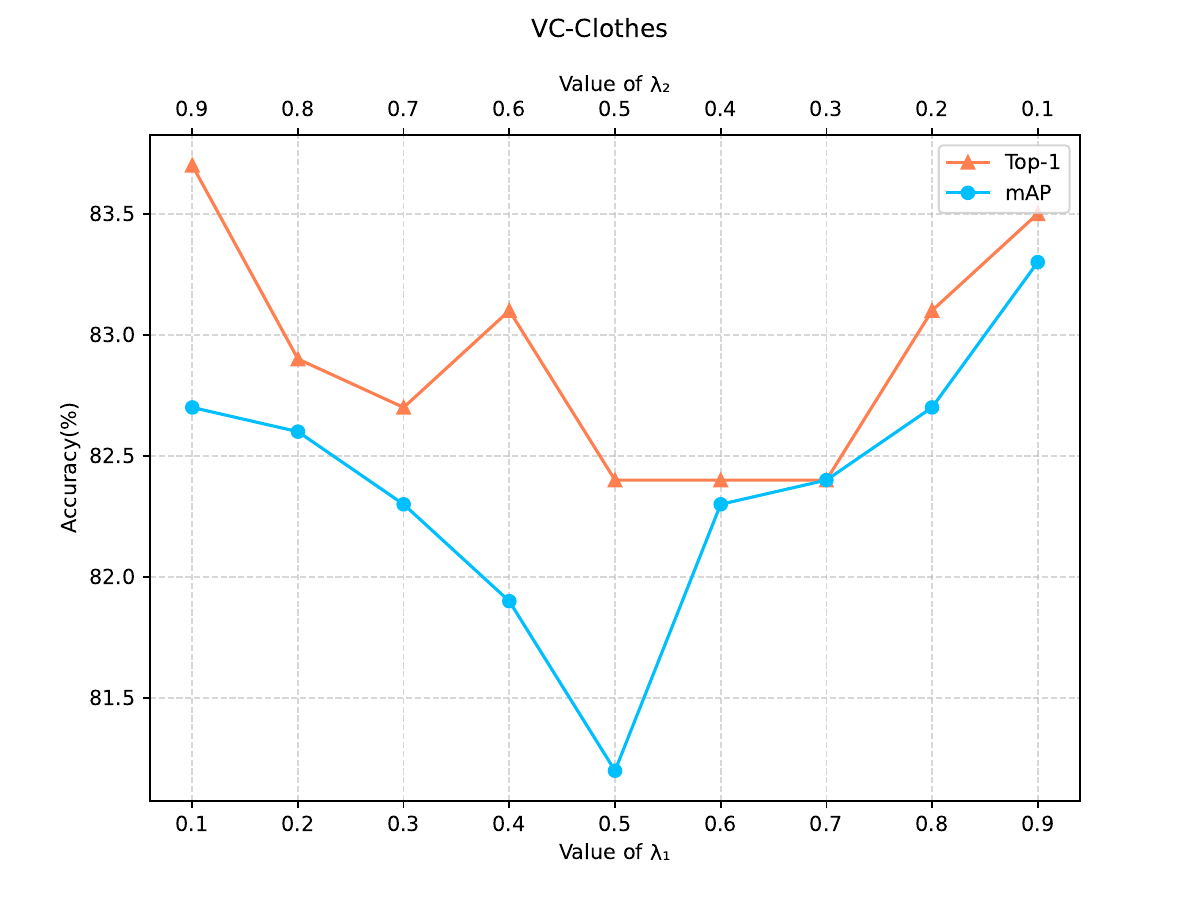}
        \caption{VC-Clothes}
        \label{fig:sub2}
    \end{subfigure}
    \caption{Impact of the hyper-parameter $\lambda_{1}$ and $\lambda_{2}$.}
    \label{fig3}
\end{figure}

During the training process, a dual-stage optimization strategy was implemented. In the initial stage, the model exclusively relied on Identity Loss to identify pedestrians in the same clothing, thereby acquiring the fundamental skill of re-recognizing individuals after they change their attire. Entering the second stage, Triplet Loss was integrated alongside Identity Loss to further aid the model in learning to identify more challenging pedestrians. To discover the optimal balance between the loss functions, we adjusted the weights of Identity Loss ($\lambda_1$) and Triplet Loss ($\lambda_2$), as illustrated in Figure ~\ref{fig3}. In the course of the experiments, the model reached its peak performance when the hyperparameter $\lambda_1$ was set to 0.1, thereby making $\lambda_2$ 0.9. This outcome suggests that, in the specific context of this experiment, Triplet Loss played a pivotal role in enhancing the performance of the task of re-identifying pedestrians with clothing changes. The potential reason for this is that Identity Loss had already been thoroughly optimized in the first stage, reaching a saturation point. The introduction of Triplet Loss provided the model with new learning impetus and direction, enabling it to excel in more complex scenarios.

\subsection{Qualitative Analysis through Visualization}
\begin{figure}[htbp]
    \centering
    \begin{subfigure}{0.95\linewidth}
        \includegraphics[width=\linewidth]{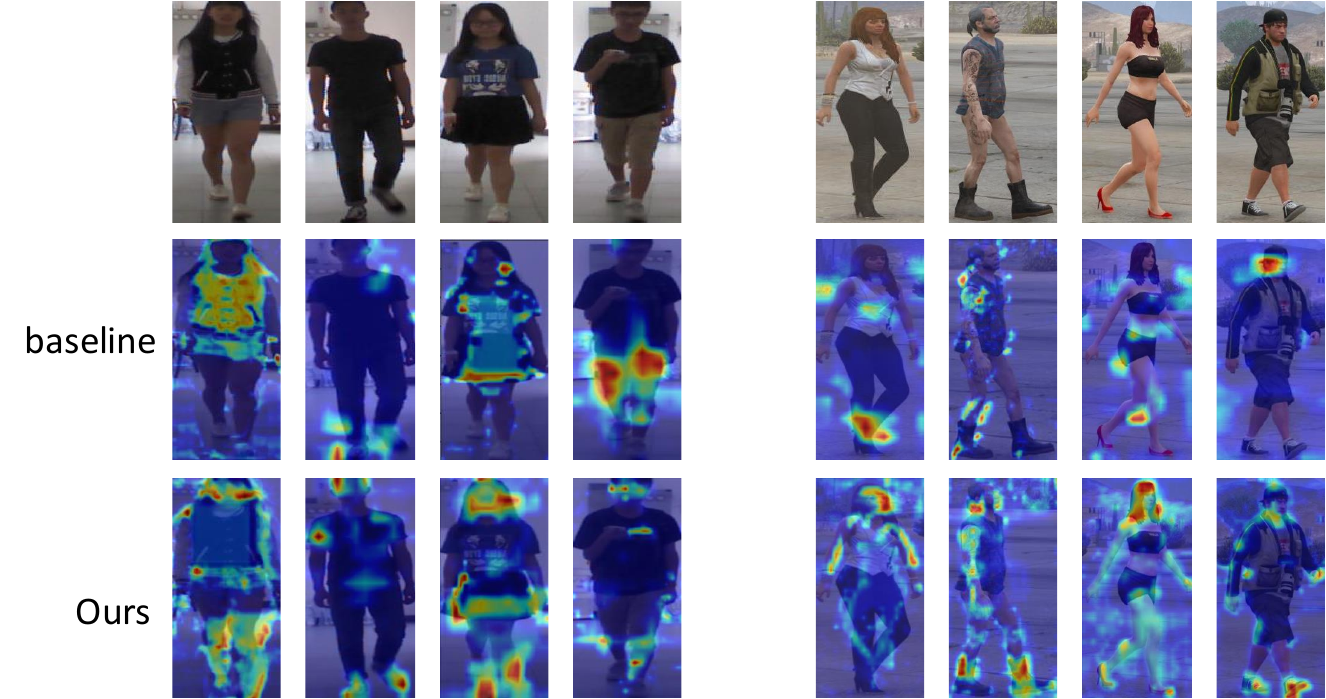}
    \end{subfigure}
    \caption{Visualization of heatmaps on PRCC (left) and VC-Clothes (right). Highlighted areas represent regions of heightened focus by the model.}\label{fig4}
\end{figure}

In our research, as demonstrated in Figure ~\ref{fig4}, we utilized heatmaps to illustrate the pronounced differences between our method and baseline approach in the CC-ReID task. Our method distinctly focuses on disentangling key identity features of pedestrians, such as the face, arms, and legs, thereby exhibiting intense attention to these crucial areas. Conversely, baseline methods scatter their focus on non-essential areas like clothing, a trend especially noticeable in the PRCC dataset, leading to reduced accuracy. This juxtaposition clearly underscores the superiority of our approach in tackling the complexities inherent in the CC-ReID task.

\section{Conclusion}
In this study, we propose an efficient and concise Clothes-Changing Person Re-Identification (CC-ReID) method. The essence of this method lies in extracting disentangled clothing-irrelevant features, such as the face, arms, and legs, by leveraging semantic segmentation maps obtained from pedestrian parsing. This process effectively excludes variable elements like clothing. These constant features are then fed into the model for deep learning and recognition. To further enhance the model's discriminative power, we introduce a gated channel attention mechanism to delve deeper into the distinctive characteristics of the disentangled features. Moreover, extensive experiments validate the efficacy of our proposed method, demonstrating superior performance over several current state-of-the-art approaches.

\end{document}